# Threshold-Based Selection for Continuous Optimization: A Leaf-Abscission Instantiation

Nasser Khalili
*Graduate School of Management and Economics, Sharif University of Technology, Tehran, Iran.*
nasser_khalili@gsme.sharif.edu

***Abstract*—This paper formalizes threshold-based selection as an evaluation-gating architecture in which each incumbent is tested before variation and a replacement is generated and evaluated only when contextual pressure exceeds intrinsic strength. The mechanism is instantiated as Leaf Abscission Optimization (LAO), using rank-based strength, a phenological seasonal signal, diversity modulation, environmental pressure, and a base regrowth kernel. A blocked $2^4$ factorial analysis at $D$ = 10 on the CEC 2017 suite reduces the original multi-layer design to a parsimonious core: drift is harmful, while the other three auxiliary layers show no robust independent evidence of benefit. The resulting LAO-Core attains the third-best mean Friedman rank among nine optimizers at $D$ = 10, 30, and 50 under the equal $300D$ evaluation budget. A four-budget sweep shows budget-dependent relative performance, with adaptive differential-evolution baselines gaining relative advantage at larger budgets; the nine-cell dimension–budget analysis establishes neither an *FE/D*-only law nor a statistically significant dimension–budget interaction. A paired intervention shows that diversity modulation changes late-run replacement behaviour without a detectable effect on final error at the tested budget. The evidence supports LAO as a parsimonious evaluation-gating mechanism with regime-qualified competitiveness, rather than as a generally superior optimizer.**

## I. INTRODUCTION

THE canonical cycle of evolutionary computation couples generation and retention tightly: variation operators propose new candidate solutions, the objective evaluates them, and a selection step decides which of parents and offspring enter the next generation [1], [2]. In the conventional survivor-selection cycle considered here, these mechanisms determine retention after candidate generation and objective evaluation. A large class of practical optimisation problems, however, is dominated by evaluation cost rather than by the algorithmic overhead of generating or ranking candidates, and in this regime the binding constraint may be the number of exact objective evaluations incurred. An algorithm that can withhold search action — that can decide, per individual and per step, whether generating a replacement is worth an evaluation — may offer an architectural advantage in settings where exact objective evaluations dominate computational cost, although lower evaluation count does not imply lower wall-clock time or better final accuracy. Steady-state, asynchronous, and selective-evaluation methods are important exceptions that we position against in Section II.

This paper studies a mechanism that we call threshold-based selection. Every candidate carries an intrinsic strength, the population exerts an abscission pressure on each candidate, and a candidate is replaced — and only then is a replacement generated and evaluated — precisely when its pressure exceeds its strength. The distinction from conventional survivor selection is architectural rather than parametric: in a movement-first algorithm, selection controls which solutions are retained after they have been generated, whereas in a threshold-based algorithm, selection controls whether search action occurs at all. A candidate that does not cross its threshold consumes no evaluation and remains bit-identical, a property we enforce structurally rather than assert. The gate therefore precedes both variation and exact objective evaluation, which is the central architectural claim of this paper and the property that distinguishes it from surrogate-assisted methods that follow a construct-then-surrogate-then-evaluate-exactly pathway rather than the threshold-then-decide-whether-to-construct pathway used here. This ordering distinguishes the gate from parameter changes within conventional survivor selection.

The threshold rule is instantiated with the quantitative phenomenology of leaf abscission, the process by which deciduous trees shed leaves. We treat the biological literature as a source of models rather than as a metaphorical licence. Wang et al. [3] fit the logistic model F(t) = $P_1$ / [1 +

$\exp((2.2/P_3)(P_2 - t))$] to cumulative leaf-fall curves, where $P_2$ is the timing of peak fall and $P_3$ is the interval between 10% and 50% cumulative fall; Dixon [52] introduced a quantitative cumulative leaf-fall model with biologically interpretable parameters that grounds this family. This provides the biological basis for the functional form used in the seasonal pressure term; the corresponding optimization mapping is defined here, and the slope is fixed to the cited $2.2/P_3$ form so that $P_3$ retains its stated 10%-to-50% interpretation. Wang et al. [3] also report early, intermediate, and late phenological classes, and Scala and Cappellini [4] model abscission as a survival process driven by intrinsic resistance weakened by endogenous and exogenous effects, with Addicott [5] providing the foundational physiological treatment of the separation layer and hormonal regulation. This motivates a per-individual strength that depends on a species class and on relative fitness.

The third model input is environmental modulation. Taylor and Whitelaw [6] establish that environmental cues modulate the separation layer through hormonal and physiological signals; Estiarte and Peñuelas [50] show that photoperiod, temperature, and water stress can each alter the timing and proficiency of leaf senescence and fall; Kane et al. [51] provide recent evidence that abscission and senescence are distinct processes whose temporal coupling varies across species; Pautot et al. [53] characterize the abscission zone as a specialised cellular interface for programmed organ separation, and Hikosaka et al. [54] examine leaf shedding as an optimization of resource allocation under environmental constraints. This motivates the environmental term of the pressure equation; because the underlying engineering models — turbulence spectra, discrete gusts, and mean vertical profiles — are not interchangeable, the implementation distinguishes these roles explicitly rather than sampling a generic wind speed per candidate. The biological model does not provide convergence guarantees, algorithmic parameter values, or evidence of biological fidelity. The algorithm is inspired by the abscission event — the discrete separation of a leaf from its branch — and not by the prior biochemical cascade of senescence; this distinction is preserved in the algorithmic mapping.

A pre-specified component analysis led to a substantially simpler architecture than the initially considered multi-layer design: when every auxiliary component was required to justify itself in a blocked factorial analysis, drift proved harmful and the remaining optional layers showed no robust independent contribution after drift removal. We therefore retain the threshold-selection core and disable the auxiliary operators not supported by the component analysis. The principal claims are: (i) a formal specification of threshold-based selection as an architectural principle distinct from movement-first selection, with a rank-based strength rule invariant to strictly monotone transformations of objective values, a phenologically scheduled pressure term, and structurally enforced stationarity (Section III); (ii) a parsimonious instantiation (LAO) whose final specification is a threshold core with a base regrowth kernel, where optional layers are disabled after component-level analysis rather than retained by design intent (Sections IV and VII-B); (iii) an empirical characterisation of a budget-dependent performance regime that is competitive under the tested constrained budget, with changing relative standing as the evaluation horizon approaches the 10,000·D budget commonly used in CEC-style benchmarking [7] (Section VII-C); and (iv) causal behavioral evidence for the effect of diversity modulation on replacement activity, provided by 660 instrumented runs reproducing implementation-level relationships and a paired intervention (Section VII-D).

At every dimension tested, CMA-ES [8], [9] and L-SHADE [10] are equal or better on aggregate. LAO remains competitive under the tested 300·D budget, but its relative standing changes at larger budgets. We follow the current methodological consensus for algorithm comparison: the official CEC 2017 benchmark definitions and support data [7]; blocked non-parametric testing with the benchmark function as the block [18], [19]; a single post-hoc procedure with family-wise correction [20]; effect sizes on raw run values [21], [22]; and anytime analysis indexed by consumed evaluations rather than iterations [23]. The remainder of the paper is organised as follows. Section II positions threshold-based selection against selection, restart, and population-management mechanisms, with explicit attention to plant-inspired optimizers. Section III formalises the paradigm and states Definition 1. Sections IV–VI specify the algorithm, develop the state-space formulation, and detail the benchmark and statistics. Sections VII–X present results, discussion, limitations, and conclusions.

## II. RELATED WORK

The relevant literature separates into five partially overlapping families of mechanism that bear on threshold-based selection. The first family comprises survivor and replacement selection within the standard generation-evaluate-select cycle. Selection is one of the two forces — with variation — that define an evolutionary algorithm [1], [2]. In the conventional survivor-selection cycle considered here, these mechanisms determine retention after candidate generation and objective evaluation. The distinction we draw is not between strong and weak selection but between two placements of the selection decision in the algorithmic cycle: movement-first architectures cannot withhold an evaluation once the operator has proposed a candidate, whereas threshold-based architectures decide first whether the slot is to be vacated at all. Parameter-control research [11] recognises that the balance between exploration and exploitation should change over a run, and adapts operator parameters accordingly; threshold-based selection addresses a complementary question, not how strongly to move but when to act on an individual at all. The two are orthogonal — an adaptive parameter schedule could drive the pressure term of a threshold algorithm — and we treat the interaction as future work.

The second family comprises temporal and individual-level replacement control: restart, extinction, and population-size management schemes that gate search action on aggregate criteria. Restart strategies [12] re-initialise individuals or the whole population when a stagnation criterion fires, and extinction-based approaches remove a fraction of the population on schedule or on deterioration. Population-size adaptation [13], [14] modulates the number of active individuals over the run. All of these gate search action on a population-level criterion evaluated on aggregate signals such as best-so-far stagnation or elapsed evaluations. Threshold-based selection differs in granularity and timing: the decision is per incumbent and is made before candidate construction. CHC [37] already provides a pre-recombination threshold; GAVaPS [38] and ALPS [41] provide persistent individual lifetime or age; crowding [42], sharing [44], and clearing [43] provide population-context-dependent suppression; and evolution-control and surrogate-assisted methods selectively allocate exact objective evaluations [39], [40]. Aging-based methods form a sub-family structurally closest to threshold-based selection: individuals expire and are regenerated after a lifetime parameter that may be fixed, randomised, or fitness-dependent. Threshold-based selection differs from this mechanism along a different axis, replacing a clock or a fitness-only criterion with a composite per-individual condition — pressure versus strength — in which the pressure side is designable. Consequently, LAO does not claim novelty for thresholding, individual state, contextual selection, or selective evaluation separately. To the best of our knowledge, its narrower contribution is the direct per-incumbent comparison of explicit contextual pressure with explicit intrinsic strength before variation, where failure of the comparison suppresses both candidate construction and objective evaluation.

The third family comprises biologically inspired optimizers, and within this family the recent plant-inspired literature is the most direct precedent for our work. Fang and Cao's Leaf in Wind Optimization [45] models falling leaves in wind as a search mechanism, using linear translation and spiral rotation as the operator that moves candidate solutions through the search space; the algorithm was validated on the CEC 2017 suite at $D \in \{10, 30, 50, 100\}$ under a budget of $5000 \cdot D$ evaluations. LAO abstracts a different sub-event of the same biological substrate: leaf abscission, the discrete separation of a leaf from its branch, rather than leaf motion in air. Both algorithms draw inspiration from leaf-fall phenomena, but they abstract different processes — motion versus separation — and therefore implement different search mechanisms. Seasons Optimization (SO) [46] maps the tree growth cycle to multiple search operators — renew, competition, seeding, and resistance — while Enhanced Seasons Optimization (ESO) [47] extends this set with root spreading, wildfire, enhanced competition, enhanced resistance, and opposition-based learning. Several plant-inspired optimizers incorporate multiple biologically motivated operators; in the studies examined here, component-level attribution is less consistently documented than aggregate algorithmic performance. Our component analysis instead evaluates whether each additional mechanism earns its computational and conceptual complexity, and finds that for LAO three of the four auxiliary layers did not. A recent systematic review of 175 plant-inspired optimization studies [48] confirms that plant-inspired optimization is an active family rather than a marginal one. The claimed contribution lies in the particular placement of the gate and in coupling a failed threshold decision to the suppression of both candidate construction and exact evaluation.

The fourth family comprises adaptive differential evolution and hybridization with local refinement, which together provide our strongest baselines. SHADE [15], L-SHADE [10], and their lineage through JADE [16] and jDE [17] are adaptive DE variants that accumulate historical success information to steer mutation parameters, with optional archives of inferior solutions. Their architecture is movement-first with sophisticated adaptation, and their budget behaviour — their rank improving dramatically as the evaluation budget grows — is the expected signature of methods whose adaptive success-history mechanisms may become more effective as the available evaluation horizon increases. We use this expected behaviour to delimit the regime in which a simpler gating mechanism is competitive. Memetic algorithms [36] couple population-based global search with individual-level local refinement, and the competence of such hybrids is known to depend on how the local search effort is allocated. The elite-refinement component of the original LAO architecture was precisely such a hybrid layer: a candidate-replacement local search applied to the top-K slots. Its fate in our ablation — no demonstrated contribution once the drift layer with which it interacted was removed — is an instance of the allocation problem commonly encountered in memetic designs: local refinement earns its evaluations only when the global layer leaves it useful work. The factorial methodology used here, which measured the elite layer both unconditionally and conditional on drift's removal, is directly applicable to component questions in memetic designs.

The fifth and final family is methodological: benchmarking practice and statistical comparison. We follow the current consensus for algorithm comparison: the official CEC 2017 benchmark definitions and support data [7]; blocked non-parametric testing with the benchmark function as the block [18], [19]; a single post-hoc procedure with family-wise correction [20]; effect sizes on raw run values [21], [22]; and anytime analysis indexed by consumed evaluations rather than iterations [23]. Scale-free target definitions and empirical target-attainment distributions follow the practice popularised by the COCO platform [23]. The aligned rank transform [35] is used as a non-parametric check on the factorial analysis. The same protocol is used to define the inferential basis for the comparisons reported below.

## III. THRESHOLD-BASED SELECTION: A FORMAL FRAMEWORK

Before stating the framework, we record the architectural property that distinguishes threshold-based selection from existing selection mechanisms. Definition 1 (Evaluation-gated selection). A selection mechanism is evaluation-gated when its decision is made before candidate construction and determines whether an exact objective evaluation is incurred. By this definition, classical survivor selection is not evaluation-gated: it operates after candidates have been constructed and evaluated. Surrogate-assisted methods may suppress exact evaluation after candidate construction, and therefore differ from the strict form defined here. Threshold-based selection as formalised below is evaluation-gated in the strict sense: failure of the threshold test suppresses both candidate construction and exact evaluation. LAO satisfies Definition 1 by construction. The implementation identity of Section V verifies the evaluation-suppression part empirically across all 660 instrumented runs with zero violations; the candidate-construction-suppression part is verified separately by an instrumented counter that records every call to the regrowth kernel and asserts equality with the count of evaluated threshold crossings.

Let the population at step $t$ be $X_t = (x_{1,t}, \ldots, x_{N,t}) \subset \mathbb{R}^D$, with each candidate carrying auxiliary attributes: a species label $s_{i,t} \in S$ and a canopy height $z_{i,t} \in (0, 1]$. Two per-individual quantities are computed at every step. The intrinsic strength is defined in (1) below, where $r_{i,t}$ is the zero-based rank of candidate i's objective value within the population (0 = best, $N-1$ = worst) and $\ell(s)$ is the baseline strength of species class $s$.

$$LS_{i,t} = \ell(s_{i,t})\left(1 - \frac{r}{N-1}\right) \tag{1}$$

Two properties of this rule are deliberate. First, $LS$ depends on fitness only through its rank, so the rule is invariant to any strictly monotone transformation of the objective; this matters on benchmarks such as CEC 2017 whose population fitness ranges span many orders of magnitude, where the rank-based construction avoids the numerical compression observed with outlier-sensitive min–max normalisation. Second, $LS$ is bounded in $[0, \ell_{\max}]$, which keeps the threshold interpretable on a normalized decision scale; the parameterization keeps both sides of the threshold on a common [0,1]-compatible scale ($\alpha$ = 0.90, $\ell_{\max}$ = 0.90 in the shipped defaults). The environmental exposure is given in (2) below, where $\tau \in [0, 1]$ is normalised optimisation time indexed by consumed evaluations, $d_t$ is the normalised mean pairwise diversity of the population, $U_{i,t} \in [0, 1]$ is the environmental exposure of candidate $i$, $\xi_{i,t} \sim U(0,1)$ is an individual susceptibility draw, $S(\tau, s)$ is the phenological seasonal schedule of Section IV, and α, β, λ are coefficients. The seasonal term provides the primary structured component, while environmental exposure enters as an additional perturbation. The shared environmental field state $W_t$ (Section IV-C) is a single per-iteration draw that drives $U_{i,t}$ through a vertical profile, and we use "environmental exposure" for $U_{i,t}$ and "environmental field" for $W_t$ consistently throughout.

$$AP_{i,t} = clip(\alpha S(\tau, s_{i,t})\,(1 + \lambda(1 - d_t)) + \beta\, U_{i,t}\, \xi_{i,t}, 0, 1) \tag{2}$$

The threshold rule and the replacement policy are given in (3) below, where $G$ is the regrowth kernel of Section IV-D. The stationarity property — a slot that does not cross its threshold is left bit-identical — is enforced structurally in the implementation: every change to a candidate's position passes through a single replace routine, and the local-search operator (where present) proposes a candidate, evaluates it, and replaces the incumbent only on strict improvement. Without this enforcement, the claim that candidates remain stationary until shed would be unsupported. Three properties of the rule are reported separately because they have different evidential status. Invariance is enforced: $LS$ depends on the objective only through within-population ranks, so the fitness-derived part of the replace/stay decision is invariant under any strictly increasing transformation of the objective. Empirical selectivity is tested: across the 660 instrumented runs of Section VII-D the median per-iteration replacement fraction is 0.067, that is, two of thirty slots per iteration, indicating selective rather than full-population replacement. Pressure–replacement association is tested: within a run, a larger pressure–strength margin is associated with more replacement ($\rho$ = 0.47, computed over aligned per-iteration observations within the instrumented runs). This association therefore provides a direct diagnostic of whether an implementation preserves the intended threshold behaviour.

$$x_{i,t+1} = \begin{cases} G(X_t, s_{i,t}, z_{i,t}, W_t), & AP_{i,t} > LS_{i,t} \\ x_{i,t+1}, & AP_{i,t} \le LS_{i,t} \end{cases} \tag{3}$$

The paradigm makes one architectural claim: selection can control whether search action occurs, not only which solutions are retained. It does not claim that the pressure equation's particular terms are necessary — that is an empirical question this paper answers negatively for three of them — nor that the threshold confers universal performance advantage; the budget analysis of Section VII-C shows that any observed relative advantage is conditional on the tested regime. We did not identify a prior study that combines these elements in the same per-incumbent, pre-variation evaluation-gating architecture; each ingredient separately is well precedented in the literature reviewed in Section II.

## IV. THE LEAF ABSCISSION OPTIMIZATION ALGORITHM

LAO instantiates the threshold-based selection framework described in Section III. At each iteration, the environmental field is updated once and shared across the population. Candidate-level environmental exposure is obtained from this shared field through the prescribed vertical profile. Population diversity, intrinsic strength, and abscission pressure are then computed for each incumbent, and only individuals for which pressure exceeds strength are selected for replacement. These threshold-crossing individuals are regenerated using the base regrowth kernel. The final architecture selected by the component analysis in Section VII-B retains the threshold-selection core and the base regrowth kernel, while the drift, gust, vibration, and elite-refinement layers are disabled. The environmental field remains part of the pressure calculation through its contextual contribution.

The complete procedure is summarized in Algorithm 1. The number of threshold crossings and the number of evaluated replacements is distinguished explicitly: $C_t$ denotes the number of incumbents satisfying the threshold condition, whereas $S_t$ denotes the number of replacements that can be evaluated under the remaining budget. Objective-evaluation counting is performed only when the regrowth kernel is executed, yielding the evaluation identity given in Section V. When the remaining budget is insufficient to process all threshold crossings, the available replacements are selected in deterministic ascending index order. This rule affects only terminal-budget truncation and does not modify the non-terminal search trajectory.

**Algorithm 1.** Leaf-Abscission Optimization (LAO)

**Input**: Objective function f, bounds [l,u]^D, budget N_max, parameters Θ
**Output**: Best solution x_best and objective value f_best

1: **Initialize** N individuals x_i ~ U(l,u), s_i ~ U{0,1,2},
   z_i ~ U(0.2,1), i = 1,...,N
2: **Evaluate** f(x_i), i = 1,...,N
3: NFE ← N
4: x_best ← arg min_{i=1,...,N} f(x_i)
5: f_best ← f(x_best)

6: **while** NFE < N_max **do**
7:   τ_t ← (NFE − N)/(N_max − N)
8:   W_t ← EnvironmentField.step(τ_t)
9:   d_t ← normalized mean pairwise distance

10:   **for** i = 1,...,N **do**
11:     ε^env_{i,t} ~ N(0,0.05²)
12:     ξ_{i,t} ~ U(0,1)
13:     U_{i,t} ← clip(W_t · h(z_{i,t}) + ε^env_{i,t}, 0,1)
14:     LS_{i,t} ← ℓ(s_{i,t})(1 − r_{i,t}/(N−1))
15:     AP_{i,t} ← clip(αS(τ_t,s_{i,t})(1+λ(1−d_t))
16:         + βU_{i,t}ξ_{i,t}, 0,1)
17:   **end for**

18:   C_t ← #{i : AP_{i,t} > LS_{i,t}}
19:   S_t ← min(C_t, N_max − NFE)
20:   **Select** the first S_t crossing indices in ascending order

21:   **for** each selected index i **do**
22:     x_i ← Regrow(τ_t, x_best)
23:     f_i ← f(x_i)
24:     Refresh s_{i,t} and z_{i,t}
25:     NFE ← NFE + 1
26:   **end for**

27:   x_best ← arg min_{i=1,...,N} f_i
28:   f_best ← f(x_best)
29: **end while**

30: **return** x_best, f_best

## A. Pseudocode

Algorithm 1 presents the operational form of LAO. The threshold comparison $AP_i > LS_i$ is the sole selection decision: all preceding operations compute the quantities required for this comparison, while all subsequent search activity is restricted to the selected incumbents. The remaining-budget guard distinguishes the number of threshold crossings from the number of replacements that can actually be evaluated. Thus, $C_t$ measures the demand generated by the threshold rule, whereas $S_t$ determines the evaluations actually incurred during iteration $t$. The terminal truncation rule is deterministic and is used only when fewer evaluations remain than threshold crossings.

## B. Phenological Seasonal Schedule

The seasonal schedule uses Wang et al.'s logistic [3] directly, as given in (4) below, with species classes ($P_2$, $P_3$, $\ell$) as follows: early (0.30, 0.10, 0.60), intermediate (0.55, 0.15, 0.75), late (0.80, 0.12, 0.90). The logistic slope $2.2/P_3$ is the published model's; it is what makes $P_3$ the interval between 10% and 50% cumulative fall, and our unit tests assert this property numerically. The early/intermediate/late timing order follows the reported phenological classes. The numerical strength values, class-specific timing coordinates, and mapping from day-of-year to normalized optimization time are algorithmic design choices, not empirical measurements from the biological studies; Fig. 1 shows the three schedules with their derived 10%,50%,90% landmarks.

$$S(\tau, s) = \frac{1}{1 + exp((2.2/P_3(s)) \cdot (P_2(s) - \tau))} \quad (4)$$

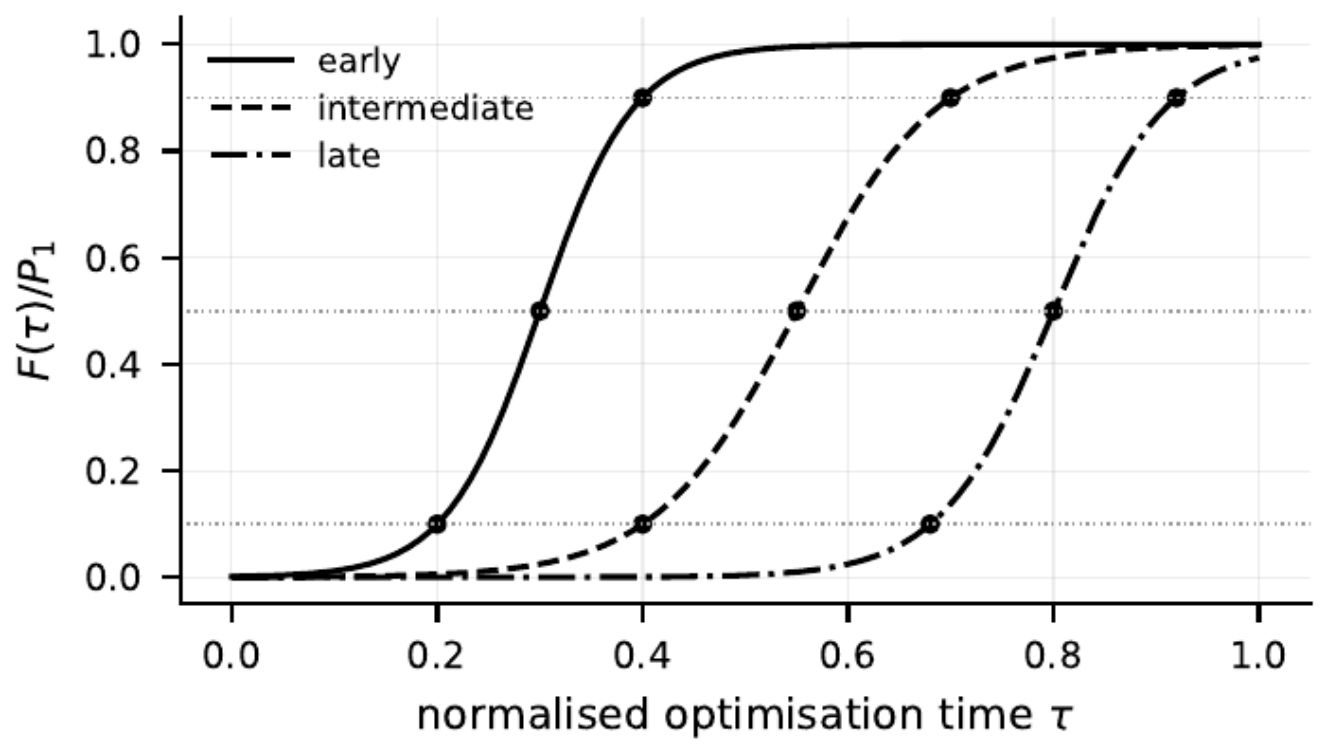

**Fig. 1.** Empirical input to the seasonal schedule. The curve is the logistic cumulative leaf-fall model of Wang et al. (2022), $F(t) = P_1/[1 + \exp((2.2/P_3)(P_2 - t))]$, for the early, intermediate, and late phenological classes they report; markers are the derived 10%, 50%, and 90% landmarks. $P_2$ and $P_3$ are empirical biological quantities; the mapping from day-of-year to normalized optimization time, and the numeric values assigned to each class, are algorithmic design choices and are not claimed to reproduce biological leaf fall.

## C. Environmental Forcing

The implementation distinguishes four environmental roles — continuous turbulence, discrete gust, mean vertical profile, and long-term marginal — because the underlying engineering models are not interchangeable [24]–[28]. Detailed equations, filter coefficients, and spectral constants are deferred to the Supplementary Material. In the final specification the exposure $U_{i,t}$ perturbs the pressure equation's $\beta$ term, but none of the optional displacement components are active.

## D. Regrowth Kernel

A threshold-crossing incumbent is regenerated using the base regrowth kernel given in (5). The kernel uses a perturbation vector $\varepsilon \sim \mathcal{N}(0, I)$, a tournament-selected parent with tournament size three, and the global best-so-far solution $x_t^{best}$. The coefficient $\eta$controls phototropic attraction, while the step scale follows $\sigma(\tau) = \sigma_0(1-\tau)^\gamma$. With probability $1 - p_{\text{local}}$, the replacement is instead sampled uniformly from the search domain. The optional wind-drift displacement and Lévy-distributed gust displacement [29] are disabled in the final specification. The species label and canopy height are redrawn after replacement; the species label represents a latent susceptibility class rather than an empirically inferred biological species, providing an algorithmic representation of heterogeneous susceptibility within the population.

$$x' = clip(x_p + \eta(x_t^{best} - x_p) + \sigma(\tau)\,\varepsilon(u_b - l_b) + optional\ terms) \quad (5)$$

## E. Schedules and Parameters

The progress variable $\tau$ is indexed by consumed evaluations rather than iteration count. This choice ensures that the internal seasonal schedule advances according to the same computational quantity used to define the evaluation budget. An iteration-based schedule would advance at different rates when algorithms or configurations perform different numbers of evaluations per iteration. The same convention is used for the vibration event schedule when that optional component is enabled. The shipped defaults are $N = 30$, $\alpha = 0.90$, $\beta = 0.20$, $\lambda = 0.50$, $p_{\text{local}} = 0.95$, $\sigma_0 = 0.15$, $\gamma = 2.0$, $\eta = 0.50$, and tournament size three. All parameters are configuration-file entries, and unknown keys are rejected during construction; the unit suite also checks that configured parameters remain reachable from the corresponding operators. None of these values was tuned on the CEC 2017 test suite; they were retained from the pre-redesign configuration. Section VII-E indicates small practical sensitivity over the tested $\pm 30\%$ perturbation range.

# V. THEORETICAL ANALYSIS

Let the algorithmic state at iteration $t$ be $S_t = (X_t, s_t, z_t, W_t, n_t)$, where $X_t$ is the population, $\boldsymbol{s}_t$ and $\boldsymbol{z}_t$ collect the individual species labels and canopy heights, $W_t$ denotes the environmental-field state, and $n_t$ is the number of objective evaluations consumed up to iteration $t$. The inclusion of $n_t$ is essential because the normalized progress variable is determined by consumed evaluations,

$$\tau_t = \frac{n_t - N}{\mathrm{N}_{\max} - N} \quad (6)$$

and therefore two otherwise identical population states at different evaluation counts can have different transition laws. The process is consequently represented as an evaluation-indexed stochastic state process whose transition kernel depends on the threshold rule, the regrowth distribution, the environmental-field dynamics, and the evaluation-dependent schedules.

The number of threshold crossings and the number of evaluations incurred at iteration $t$ are distinct. Specifically,

$$C_t = \sum_{i=1}^{N} \mathbf{1}\left(AP_{i,t} > LS_{i,t}\right) \quad (7)$$

counts all threshold crossings, whereas

$$S_t = \min(C_t, N_{max} - n_t) \quad (8)$$

counts the replacements that can actually be evaluated. Since each evaluated replacement consumes exactly one objective evaluation, the evaluation counter satisfies $n_{t+1} = n_t + S_t$, $n_0 = N = N$ and with,

$$NFE_{\text{final}} = N_0 + \sum_t S_t \quad (9)$$

Consequently,

$$\mathbb{E}[NFE_{\text{final}}] = N_0 + \sum_t \mathbb{E}[S_t] \quad (10)$$

Conditioned on the current state and before subsequent state changes, increasing the seasonal coefficient $\alpha$ or the diversity coefficient $\lambda$ weakly increases the instantaneous pressure before clipping and therefore cannot reduce the set of threshold crossings. The cumulative effect on evaluation consumption remains trajectory dependent because the remaining-budget guard couples $S_t$ to the current evaluation count. For the final LAO specification, the elite-refinement contribution is zero because that component is disabled. The implementation of the evaluation identity was verified over 660 instrumented runs with zero violations; numerical evaluation of the expectation expression using the recorded crossing series agreed with the recorded evaluation counts to a maximum absolute difference of $1.3 \times 10^{-15}$. Equation (10) is a bookkeeping relation for evaluation consumption rather than a convergence result.

Because the probability of global regeneration is state dependent and no uniform positive lower bound is established, no convergence theorem is claimed. The principal computational costs are $O(N^2 D)$ for pairwise diversity, $O(N \log N)$ for ranking, and $O(S_t D)$ for regrowth, with space complexity $O(ND)$. Thus, pairwise diversity is the dominant asymptotic overhead with respect to population size, whereas the observed runtime dependence on dimension is an empirical implementation property discussed in Section VII-E.

# VI. EXPERIMENTAL METHODOLOGY

## A. Benchmark

All experiments use the official CEC 2017 single-objective, bound-constrained test suite [7], comprising functions F1 and F3–F30; F2 is excluded in accordance with the competition specification because of numerical instability. Experiments are conducted at $D \in \{10, 30, 50\}$ over the search domain $[-100, 100]^D$. The benchmark implementation is a port of the official cec17_test_func.cpp, together with the accompanying support data, including shift vectors, rotation matrices, and shuffle vectors. The 328 support files are vendored with recorded SHA-256 hashes to preserve the exact benchmark inputs used in the experiments. The port follows the behaviour of the reference implementation where the executable code differs from the accompanying technical report, including the documented Schaffer F7 buffer handling and the inactive non-continuous rounding code in F8. This choice preserves compatibility with the published implementation rather than introducing changes based on the report description.

The implementation was validated against an unmodified binary compiled from the official reference source using 1,350 deterministic probe points. The same 64-bit linear congruential generator was used in the C and Python validation paths so that both implementations received bit-identical inputs. Across these probes, the port achieved a maximum relative error of $2.1 \times 10^{-14}$ with respect to the reference implementation.

Additional structural checks confirmed $f(o) = 100i$ for F1, F3–F8, and F10–F30, and confirmed the reference-code optimum of F9 at $z = 1$, consistent with the absence of the commonly applied $+1$ correction in the reference Levy implementation.

### B. Algorithms and Budget

Nine algorithms are compared: LAO (the final specification of Section IV), PSO (Clerc constriction [30]), GA (tournament selection, SBX with $\eta_c = 20$, and polynomial mutation with $\eta_m = 20$), DE (rand/1/bin with $F = 0.5$ and $CR = 0.9$[31]), GWO [32], WOA [33], CMA-ES [8], [9] (pycma; internal termination disabled), SHADE [15], and L-SHADE [10]. SHADE and L-SHADE use published control parameters together with the documented study-specific budget adjustment. The exact configuration of every algorithm is recorded in its run-level provenance record, and Supplementary Table S1 reports, for each baseline, the canonical setting, study setting, modification, reason, and consequence.

Three implementation decisions require explicit disclosure because they modify the canonical baselines rather than reproduce them. First, for cross-algorithm budget alignment, the coefficient schedules of GWO and WOA were reparameterized as functions of consumed evaluations rather than iteration count. These configurations should therefore be interpreted as budget-aligned study settings rather than exact reproductions of the canonical iteration schedules. Second, the internal termination criteria of CMA-ES are disabled so that the common evaluation budget, rather than an algorithm-specific stopping rule, terminates each run. Its initial mean is also sampled uniformly from the search box rather than fixed at the centre to harmonize population initialization across algorithms. This modification removes initialization and stopping differences for protocol comparability, but may alter CMA-ES performance relative to its default configuration. Third, the SHADE/L-SHADE population cap described below is applied symmetrically to both methods, changes only the population size, and is recorded in the corresponding provenance records. Cross-algorithm comparisons should therefore be interpreted as comparisons under a common evaluation-budget protocol, not as exact reproductions of the canonical default configurations of every baseline.

Every algorithm receives exactly the same number of objective evaluations for a given (function, dimension) condition. This constraint is enforced by a single evaluation counter placed before the objective-function call; an attempted call that would exceed the budget raises an exception. No algorithm can therefore overspend, no partial evaluation is counted, and the reported NFE is the value of this counter rather than the optimizer's internal bookkeeping. Control parameters are not suite-tuned; the protocol-level modifications described in this section are treated separately from parameter tuning. For baselines whose population size would otherwise prevent at least 50 generations under the reduced budget, the population is capped as documented in Table S1.

The headline comparison uses $\mathrm{MaxFEs} = 300D$, corresponding to 3,000, 9,000, and 15,000 evaluations at $D = 10,30,50$, respectively. This budget is approximately 3% of the $10{,}000D$ reference budget commonly used in CEC-style benchmarking [7] and was selected because of the measured computational cost of the complete experimental campaign. The same budget convention is stated wherever the corresponding results are reported, while the budget-sensitivity experiment in Section VII-C spans $300D$ to $10{,}000D$. Results obtained at $300D$ are therefore not compared with published CEC 2017 competition results and are not presented as such. Equal NFE ensures an equal objective-evaluation budget, but does not imply identical initial population sizes or generation counts across algorithms; this distinction is considered in the Limitations.

### C. Protocol, Runs and Provenance

Thirty independent runs are performed for each (algorithm, function, dimension) cell, with seeds 0–29 shared across algorithms. The headline comparison therefore comprises $29 \times 9 \times 30 \times 3 = 23{,}490$ runs. The ablation, budget-sweep, mechanism, intervention, and robustness campaigns are conducted separately and tracked with their own provenance records; their run counts are reported in the Supplementary Material. Every experiment is stored as an append-only provenance record, and resumed executions cannot mix incompatible protocols. As a reproducibility check, 30 stored runs sampled from the recorded experiments were re-executed using their saved protocols and seeds under the recorded software environment, yielding zero mismatches.

### D. Statistics

The inferential unit is the benchmark function ($n = 29$ per dimension), whereas effect-size distributions are computed from the 30 raw runs within each cell. Descriptive summaries are obtained by aggregation across functions using the median or mean, as specified for each analysis. For the primary comparison, we use Friedman's test with the benchmark function as the block and algorithms as treatments, evaluated using the Iman–Davenport statistic [34]. Pairwise differences are assessed with Wilcoxon signed-rank tests on per-function medians, with Holm step-down correction across the 36 prespecified pairs [19], [20]. Effect sizes are reported as Vargha–Delaney $A_{12}$ and Cliff's $\delta$, computed from the 30 raw runs and subsequently aggregated per function [21], [22]. Percentile bootstrap intervals use 10,000 resamples for medians, ranks, and evaluations-to-target.

Target levels are defined on a scale-free basis as the remaining error relative to the median error of the initial population on each function. This reference depends on the problem and initialization rather than on the subsequent behaviour of any particular algorithm; target levels range from $10^{-1}$ to $10^{-8}$. Anytime quantities are indexed by consumed objective evaluations.

The ablation analysis uses block-level factorial contrasts. For each function, the 16 variants are ranked within the block, and the main effect of component $c$ is defined as the contrast between the mean ranks of the configurations with $c$ active and inactive. The resulting 29 function-level contrasts are analysed using Wilcoxon signed-rank tests with Holm correction. A run-level analysis that retains all 30 runs within each block, together with an aligned-rank-transform check [35], is reported as a secondary analysis. Nemenyi comparisons are not reported alongside Holm correction.

A pooled test over all (function, variant) cells is not used for inference because it treats the 29 blocks of 16 dependent cells as independent observations. Section VII-B reports this analysis only to illustrate the consequences of ignoring the block structure.

## VII. RESULTS

### A. *Overall Ranking Under the Equal-Budget Protocol*

Table I and Fig. 2 report the mean Friedman rank of the nine algorithms across the 29 functions at each tested dimension. The comparison comprises 7,830 runs per dimension. The Friedman test is significant at all three dimensions ($\chi^2 = 117.5, 104.5, 86.6$ at $D = 10,30,50$, respectively; all $p < 10^{-18}$), with corresponding Iman–Davenport statistics of $F = 28.7, 22.9, 16.7$. LAO has the third-best mean rank at all three dimensions. At $D = 10$, LAO is significantly better than GWO and WOA and is not significantly different from DE or PSO. At $D = 30$, LAO remains significantly better than GWO and WOA and is also significantly better than SHADE ($A_{12} = 0.85$, 26/29 functions), while CMA-ES is significantly better than LAO ($p_{\text{Holm}} = 0.045, A_{12} = 0.27$). At $D = 50$, LAO is significantly better than GWO and WOA and is not significantly different from CMA-ES or L-SHADE. The pairwise comparison against GA at $D = 30 (A_{12} = 0.83$, 25/29 functions, $p_{\text{Holm}} = 0.22)$ and PSO at $D = 50 (A_{12} = 0.76$, $p_{\text{Holm}} = 0.061)$ illustrates that large effect sizes do not necessarily yield statistical significance after family-wise correction.

TABLE I
Mean rank under the 300$D$ evaluation budget. Values are averaged across the 29 CEC 2017 functions; brackets denote 95% bootstrap intervals. Lower rank indicates better performance. Pairwise significance is assessed separately using the prespecified Holm-adjusted tests

| Rank | D = 10 | D = 30 | D = 50 |
|---|---|---|---|
| 1 | L-SHADE 2.14 [1.83, 2.45] | CMA-ES 2.17 [1.55, 2.90] | CMA-ES 2.55 [1.79, 3.45] |
| 2 | CMA-ES 2.59 [1.97, 3.31] | L-SHADE 2.90 [2.41, 3.45] | L-SHADE 3.03 [2.52, 3.59] |
| 3 | LAO 4.07 [3.24, 4.93] | LAO 3.41 [2.72, 4.17] | LAO 3.07 [2.45, 3.76] |
| 4 | DE 4.17 [3.21, 5.17] | PSO 4.90 [4.21, 5.62] | PSO 5.31 [4.45, 6.10] |
| 5 | PSO 4.83 [4.28, 5.38] | DE 5.03 [4.07, 5.97] | SHADE 5.66 [5.10, 6.21] |
| 6–9 | SHADE 5.97; WOA 6.45; GWO 7.03; GA 7.76 | GWO 5.93; SHADE 6.41; WOA 6.79; GA 7.45 | DE 5.69; GWO 6.03; GA 6.66; WOA 7.00 |

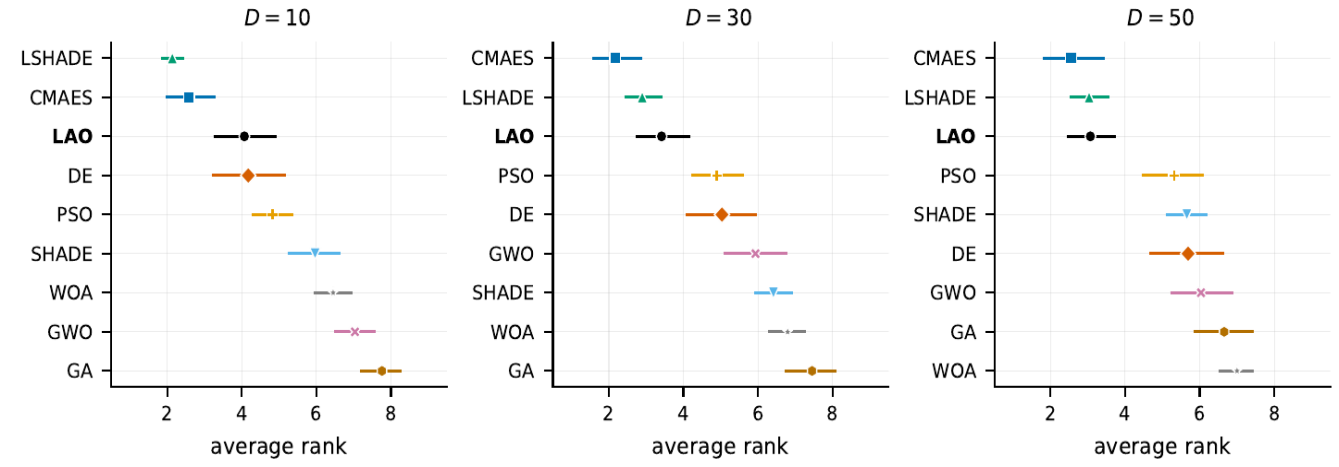

**Fig. 2.** Mean Friedman rank across the 29 CEC 2017 functions at $D = 10,30,50$. Error bars show 95% percentile bootstrap intervals obtained by resampling functions. Lower rank indicates better performance.

Fig. 3 shows the mean rank by CEC 2017 function class. LAO performs particularly well on composition functions, with a mean rank of 3.0 at $D = 10$, and on simple multimodal functions, with mean ranks of 2.57 at $D = 10$and 2.29 at $D = 30$. Its relative performance is weaker on hybrid functions, although the gap decreases with dimension, with mean ranks of 6.4, 4.6, and 3.7 at $D = 10,30,$and 50, respectively. This pattern is consistent with a possible limitation of a population-shared seasonal signal on hybrid functions, where different coordinate blocks are governed by different component functions; however, the present experiments do not isolate this mechanism causally, so the explanation remains a plausible interpretation rather than an established cause. Fig. 4 shows the anytime performance, while Fig. 5 reports the corresponding scale-free target-attainment results. At the coarsest normalized target ($10^{-1}$ of the initial error), the pooled success rate at $D = 10$is 0.785 for LAO, 0.807 for L-SHADE, and 0.761 for CMA-ES, indicating that LAO remains competitive in the early-descent regime. At the finest target ($10^{-8}$), the corresponding LAO value is 0.010, compared with 0.169 for CMA-ES, indicating a clear advantage for the adaptive baselines in the late-refinement regime.

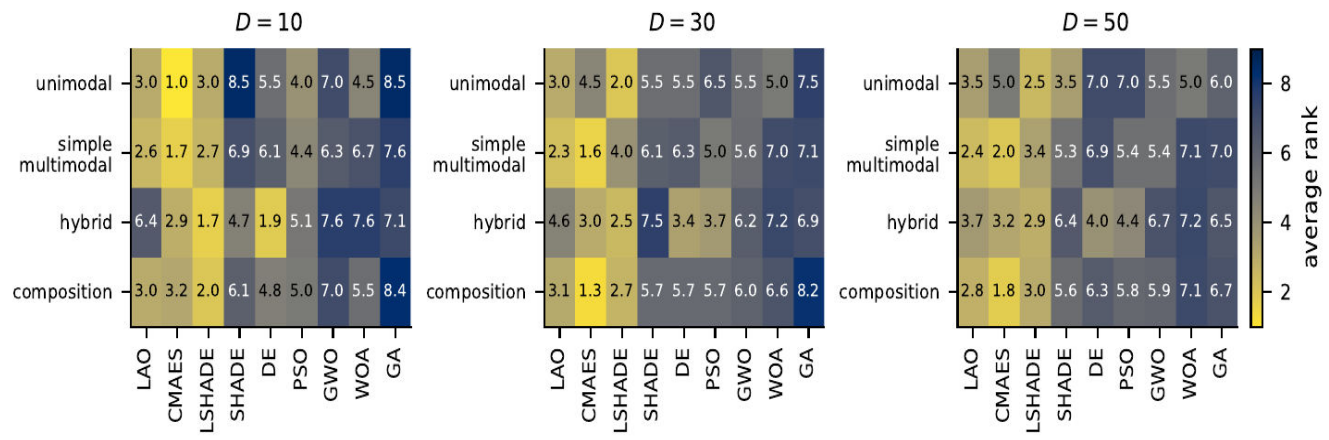

**Fig. 3.** Mean rank by CEC 2017 function class and dimension. Lower rank indicates better performance.

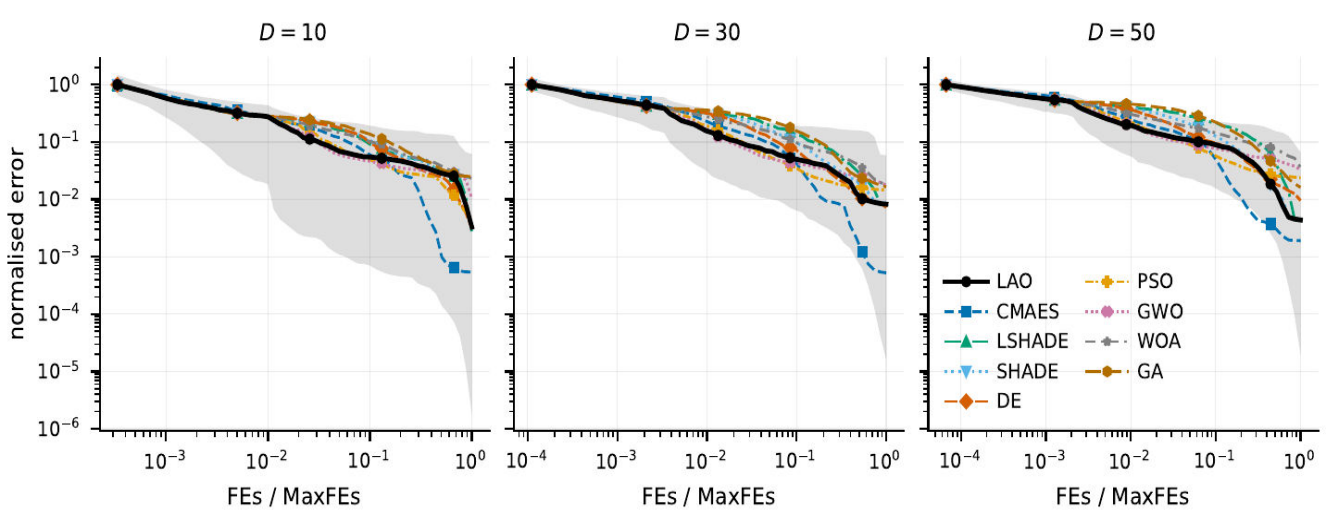

**Fig. 4.** Anytime performance as a function of the fraction of the evaluation budget consumed. Error is normalized for each function by the median initial-population error and aggregated across the 29 functions and 30 runs.

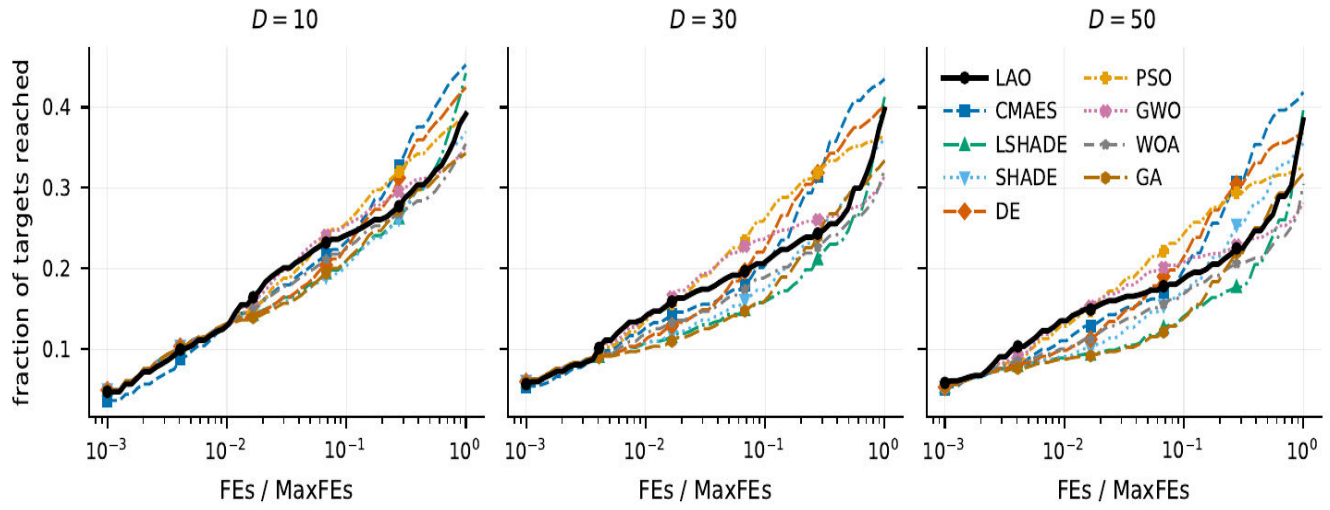

**Fig. 5.** Empirical target-attainment distributions for scale-free target levels from $10^{-1}$to $10^{-8}$. A target level $\epsilon$ is reached when the remaining error falls below $\epsilon$times the median initial-population error for the corresponding function.

Fig. 6 summarizes the Vargha–Delaney $A_{12}$comparisons between LAO and the nine baselines. Values are computed from the 30 raw runs for each function and then aggregated across the 29 functions. An $A_{12} > 0.5$favours LAO, whereas values below 0.5 favour the competing algorithm. The effect-size patterns are broadly consistent with the mean-rank results, but they should be interpreted together with the corresponding inferential tests. Moreover, Fig. 7 provides the complementary per-function comparison. Each function is classified as a significant LAO win, significant loss, or no significant difference according to a paired Wilcoxon signed-rank test on the 30 seed-matched runs, with Holm correction applied within the corresponding family of 29 comparisons. The figure shows how the aggregate results are distributed across individual benchmark functions rather than relying solely on the mean rank.

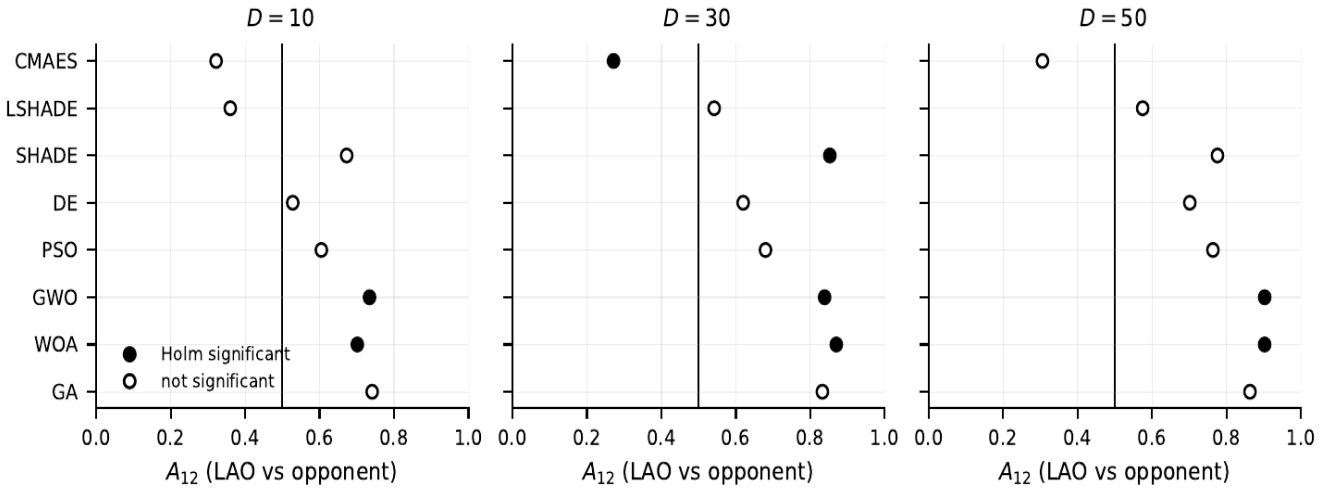

**Fig. 6.** Vargha–Delaney $A_{12}$ effect sizes for LAO against each baseline, computed from 30 raw runs per function and aggregated across the 29 CEC 2017 functions. Values above 0.5 favour LAO.

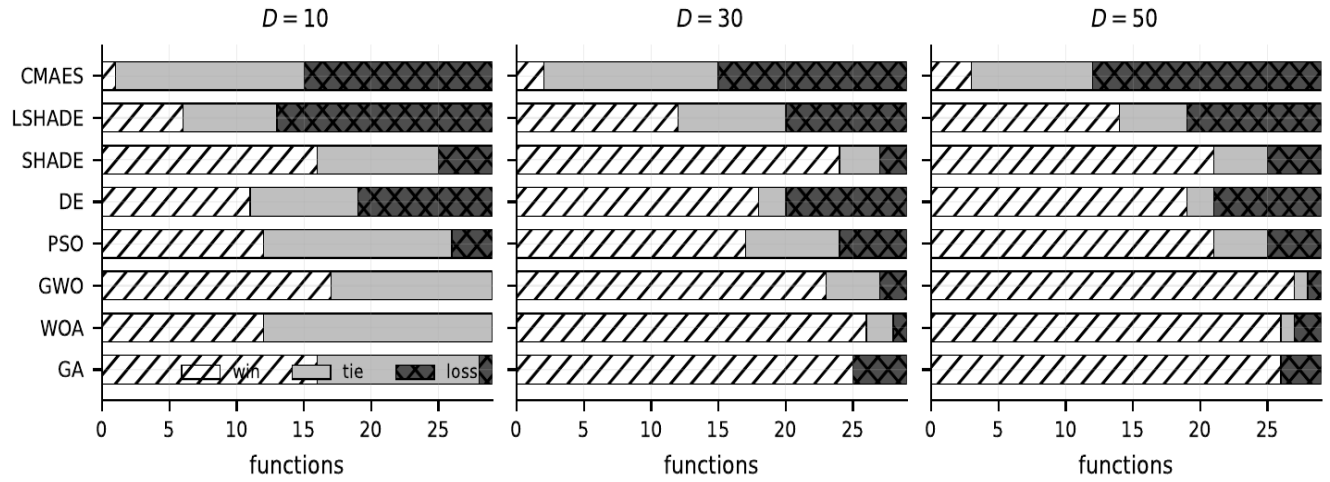

**Fig. 7.** Per-function paired comparisons of LAO against each baseline. Functions are classified as significant wins, significant losses, or no significant difference using paired Wilcoxon signed-rank tests on 30 seed-matched runs with Holm correction within each set of 29 comparisons.

### *B. The Ablation That Selected the Architecture*

The full $2^4$factorial ablation of the four auxiliary components—drift, gust, vibration, and elite refinement—comprises 16 variants evaluated on 29 functions with 30 runs per cell, for a total of 13,920 runs. The analysis uses block-level factorial contrasts as described in Section VI-D, with the benchmark function as the unit of inference. Table II reports the resulting main effects, and Fig. 8 visualizes the corresponding component contrasts and their conditional effects with drift held off.

The drift effect is supported consistently across the three complementary analyses. The run-level analysis gives a median contrast of $+1.5$, with a 95% interval of $[1.25, 1.75]$and $p_{\text{Holm}} < 10^{-4}$; the aligned-rank-transform analysis gives $F = 26.5$with $p < 10^{-4}$; and the block-level contrast also excludes zero. As shown in Fig. 8, the drift component is the only auxiliary component whose main effect is clearly detrimental under the full factorial design. In all three analyses, activating drift is associated with a deterioration in rank.

The elite component illustrates why the factorial structure is important. When all 16 configurations are considered together, elite refinement appears beneficial at the run level, with a median contrast of $-1.0$and $p_{\text{Holm}} = 7 \times 10^{-4}$. However, the drift-by-elite interaction is significant ($-4.00$, 95% interval $[-, 6.50 - 2.00]$, $p_{\text{Holm}} = 0.007$). The right panel of Fig. 8 shows that, once drift is disabled, the apparent elite benefit disappears. The corresponding sub-factorial comparison provides no evidence of an independent elite effect at either the block level ($p_{\text{Holm}} = 0.58$) or the run level ($p_{\text{Holm}} = 0.41$). The run-level result is therefore consistent with elite refinement partly offsetting the degradation associated with drift rather than providing an independent performance benefit.

Table II
Factorial main effects on within-function rank. Negative contrasts indicate better rank when the corresponding component is enabled. The inferential unit is the benchmark function ($n = 29$).

| Component | Median contrast | 95% CI | $p_{\text{Holm}}$ | Verdict |
|---|---|---|---|---|
| drift | +2.00 | [1.00, 3.00] | 0.0024 | significantly degrades rank |
| gust | −0.25 | [−1.00, 0.00] | 0.45 | not established |
| vibration | −0.50 | [−0.75, 0.00] | 0.18 | not established |
| elite | 0.00 | [−5.00, 3.75] | 0.58 | not established |

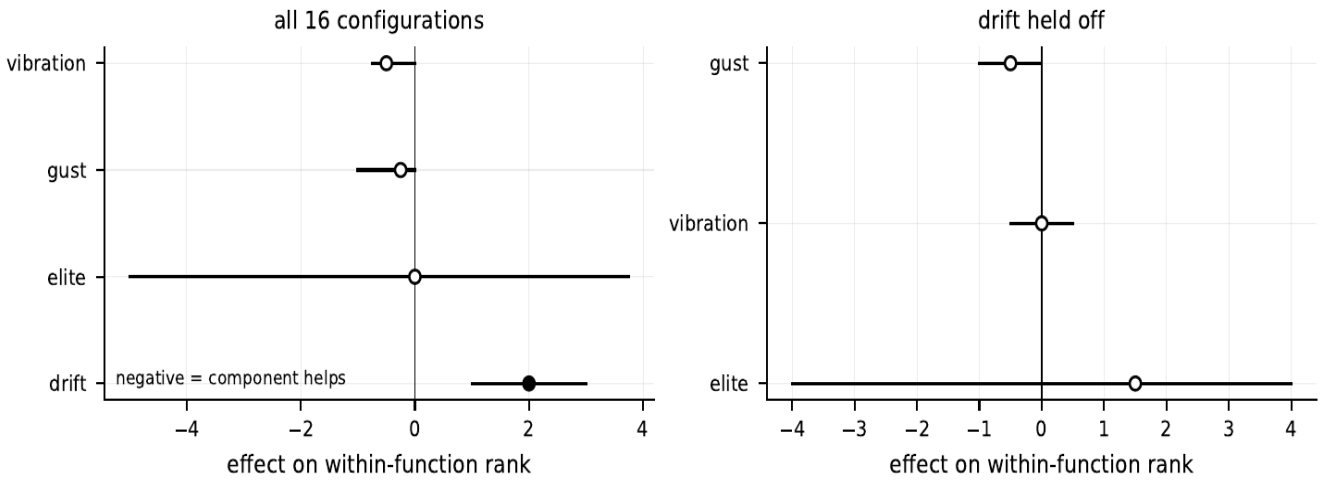

**Fig. 8.** Main and conditional component effects from the $2^4$factorial ablation. Points show the median block-level contrast across the 29 benchmark functions, with 95% bootstrap intervals obtained by resampling functions. Negative values indicate an improvement in rank when the component is enabled. Left: main effects across all configurations. Right: effects with drift disabled.

This analysis motivates the parsimonious architecture used in the remainder of the study. Under the final protocol at $D = 30$, the three-way comparison among LAO-Core, LAO-Gust, and LAO-Full comprises 2,610 runs and yields $p = 0.186$, with no pairwise difference significant after Holm correction; their mean ranks are 2.00, 1.59, and 2.41, respectively. A second comparison across the eight drift-free candidates at $D = 10$also finds no significant overall difference ($\chi^2 = 5.86, p = 0.556$). LAO-Core is therefore retained by parsimony: it is the simplest candidate not shown to be significantly inferior in the tested comparisons. Removing the auxiliary components also reduces the descriptive mean rank at $D = 30$from 3.90 for the original five-layer design to 3.41 for the parsimonious configuration, using the same protocol, seeds, and retained dataset.

### C. *Budget-Dependent Performance Regime*

A four-budget sweep on a pre-specified stratified subset of 11 functions considers $300D$, $1000D$, $3000D$, and the $10{,}000D$budget used in the CEC-style reference protocol. LAO's mean rank changes from 3.73 to 4.64, 5.27, and 5.00 across these budgets, whereas SHADE improves from 6.45 to 5.45, 3.09, and 1.68. The full 29-function confirmation at $1000D$is complete for $D = 10$and $D = 30$: LAO changes from rank 4.07 to 4.28 at $D = 10$and from 3.41 to 3.83 at $D = 30$, remaining third at $D = 30$. The corresponding $D = 50$campaign is incomplete and is therefore not analysed as a full-suite result.

A separate fixed-budget grid at $FE \in \{3000, 9000, 15000\}$and $D \in \{10, 30, 50\}$was reconstructed directly from the raw run records. For each function and budget–dimension cell, the median of the 30 runs was computed for each algorithm, all nine algorithms were then ranked within the function, and the resulting ranks were averaged across the 11 functions. At the three cells with $FE/D = 300$, LAO has descriptive mean ranks of 3.727, 3.455, and 3.000 for $D = 10,30,$and 50, respectively. These differences are not statistically significant in the blocked Friedman comparison ($p = 0.552$), and none of the Holm-adjusted pairwise comparisons is significant. A joint descriptive model, rank ~ $D + \log(FE) + D\log(FE)$, has lower leave-one-cell-out error than the $FE/D$-only model, rank ~ $\log(FE/D)$, but the interaction term is not significant (Holm $p = 1.000$). Thus, the present data do not establish either an $FE/D$-only law or a statistically significant dimension-by-budget interaction. The supported conclusion is limited to budget-dependent variation across the sampled grid.

Fig. 9 summarizes the ranking changes across the four-budget sweep, while Fig. 10 presents the corrected dimension–budget grid and the corresponding evaluation-utilization ratio. Fig. 11 shows the associated optimization trajectories against consumed evaluations at $D = 10$. Taken together, these analyses indicate that LAO remains competitive in the constrained-budget regime, whereas adaptive differential-evolution methods gain relative advantage as the evaluation horizon increases. These observations do not establish general superiority or an exact transition point.

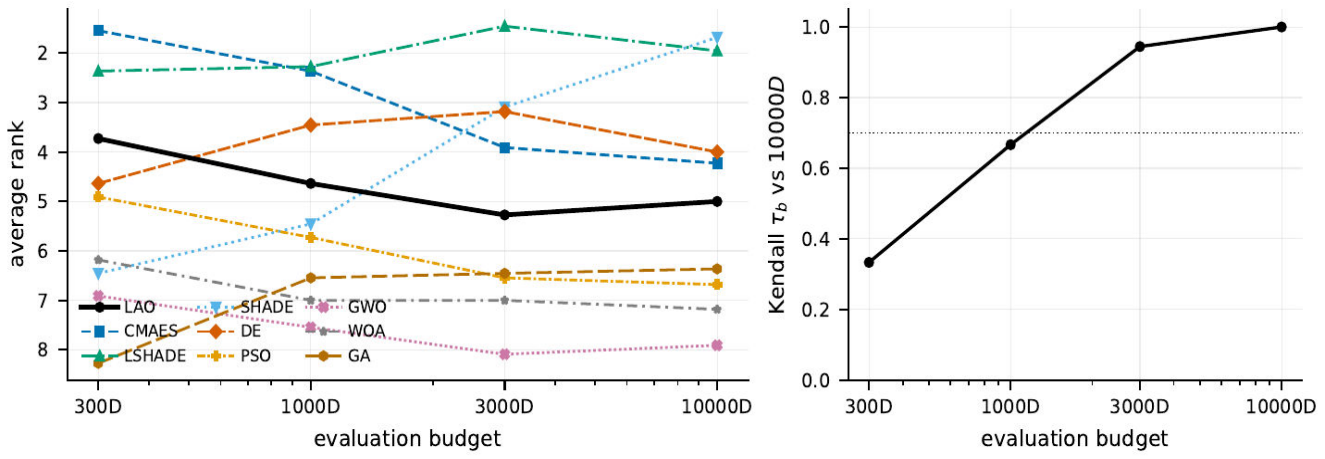

**Fig. 9.** Mean algorithm rank across four evaluation budgets on the pre-specified stratified subset of 11 CEC 2017 functions. Left: mean rank as a function of budget. Right: Kendall $\tau_b$between the ranking at each budget and the ranking at $10{,}000D$. Lower rank indicates better performance.

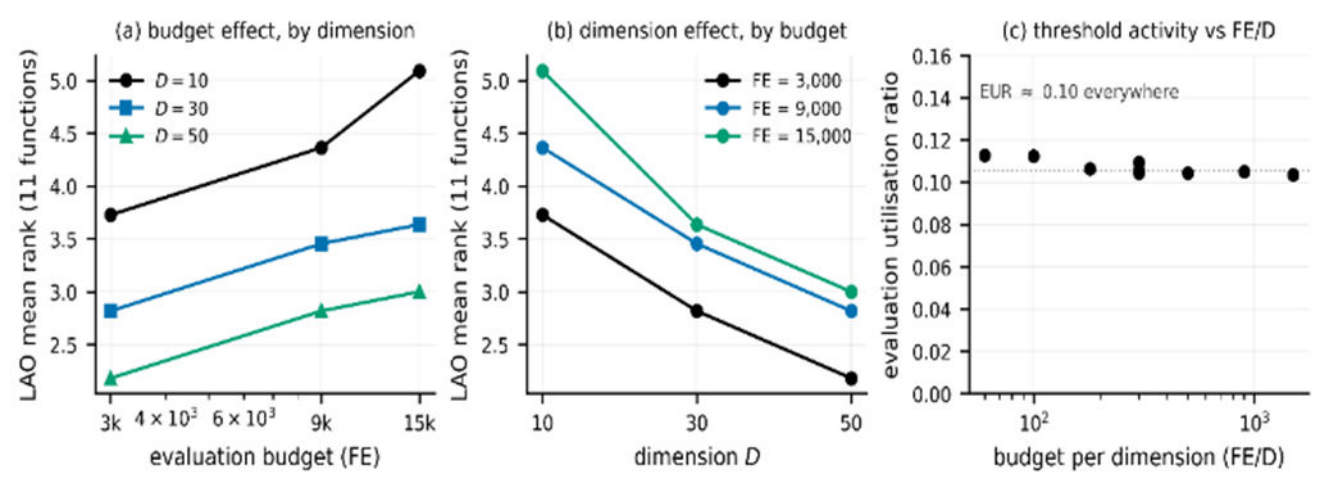

**Fig. 10.** Corrected dimension–budget grid on the 11 pre-specified stratified functions. (a) LAO mean within-function rank versus absolute evaluation budget for each dimension; (b) corresponding descriptive variation across dimensions at fixed budgets; (c) evaluation-utilization ratio across the grid. Error bars denote 95% confidence intervals over functions.

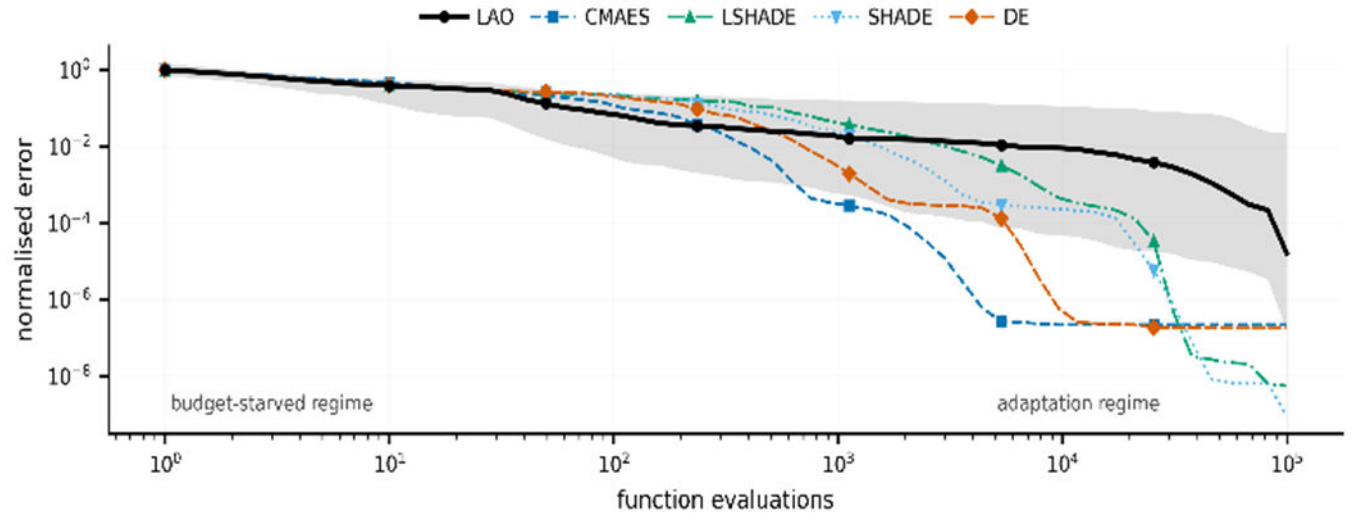

**Fig. 11.** Median normalized error versus consumed evaluations at $D = 10$on the 11 pre-specified stratified CEC 2017 functions under the $10{,}000D$budget. Error is normalized for each function by the median initial-population error and aggregated across algorithms and runs.

### D. Mechanism Evidence and a Causal Intervention

Fig. 12 summarizes 330 instrumented runs at $D = 30$(11 pre-specified functions × 30 seeds) on a common evaluation grid. The figure shows the median pressure, intrinsic strength, replacement fraction, and population diversity, together with interquartile ranges. The instrumented runs provide several implementation and behavioural checks. First, the replacement count matches the pressure–strength crossing count exactly ($\rho = 1.00$, bootstrap CI $[1.00, 1.00]$), confirming the correspondence between the threshold decision and replacement activity. Second, the pressure–strength margin is positively associated with the replacement fraction ($\rho = 0.47$), based on aligned per-iteration observations with a non-parametric bootstrap of 10,000 resamples. Third, displacement magnitude decreases over the run ($\rho = -0.21$), consistent with the prescribed decaying step schedule. The median per-iteration replacement fraction is 0.067, corresponding to approximately two replacements among 30 individuals, confirming that the mechanism is selective rather than full-population replacement.

The evaluation-count identity in Section V is also reproduced on these instrumented runs: Equation (7) holds with zero violations across all 660 runs, and the empirical evaluation of Equation (8) differs from the recorded evaluation count by at most $1.3 \times 10^{-15}$. These checks establish consistency between the implementation and the formal evaluation accounting, rather than evidence of superior optimization performance. Per-function head-to-head outcomes provide an additional descriptive view. Because seeds are shared across algorithms, the corresponding run-level comparisons are seed matched; the complete $9 \times 9$win/no-difference/loss matrix is therefore reported descriptively in the Supplementary Material. At $D = 30$, LAO has more wins than losses against GA, GWO, SHADE, and WOA, while the record against PSO and DE is more balanced. Against CMA-ES, the balance favours CMA-ES at $D = 30$and is approximately even at $D = 50$, consistent with the aggregate results in Table I.

Fig. 12 also reveals a negative association between diversity and pressure ($\rho = -0.89$). This correlation cannot by itself be interpreted as evidence for the diversity term because the term is explicitly included in the pressure equation. We therefore examine the term using a paired intervention in which the diversity weight is set to 0.5 in the control condition and 0.0 in the treatment condition. The intervention uses identical seeds, 11 functions, and 30 runs per function at $D = 10$and $D = 30$, yielding 660 matched run pairs per dimension and 1,320 runs in total. Paired Wilcoxon signed-rank tests are used for both behavioural and performance outcomes.

Prediction P1, that the control condition exhibits a negative diversity–pressure correlation, is supported ($-0.90/-0.89$). Prediction P2, that removing the diversity term eliminates this correlation, is not supported ($-0.91/-0.92$). Prediction P3, that removing the term reduces late-run replacement activity, is supported: the replacement measure changes from 0.567 to 0.400, with paired Wilcoxon $p \approx 10^{-55}$at both dimensions. Prediction P4, concerning final error, is not supported ($p = 0.31$ at $D = 10$and $p = 0.54$at $D = 30$; error ratio approximately 1.0).

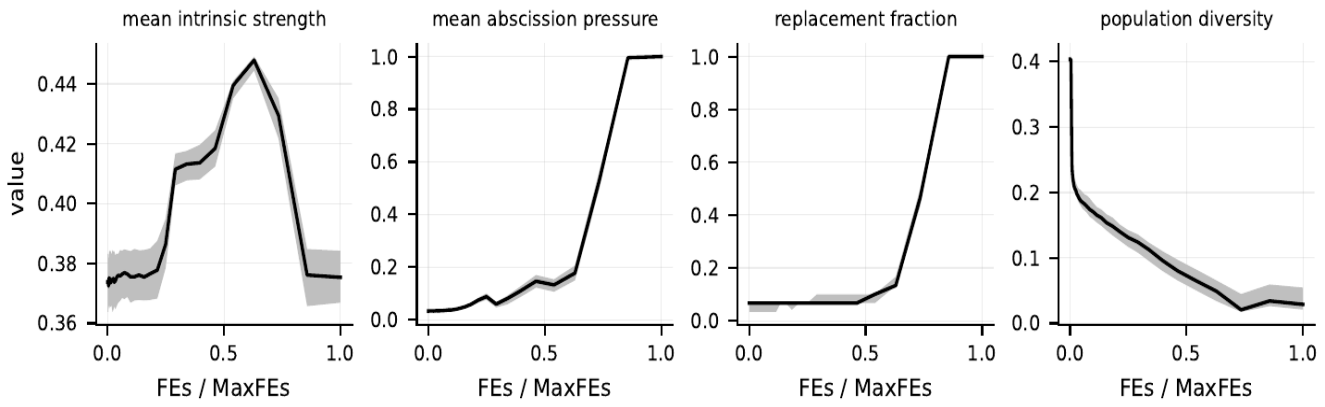

**Fig. 12.** Threshold dynamics at $D = 30$across 330 instrumented runs. Solid lines show medians and shaded bands show interquartile ranges for pressure, strength, replacement fraction, and population diversity. The replacement count matches the threshold-crossing count, and the pressure–strength margin is positively associated with replacement activity. The diversity–pressure correlation reflects shared temporal variation and is examined separately through the paired intervention.

The failure of P2 indicates that the observed diversity–pressure correlation is largely driven by their shared temporal trends rather than by the direct action of the diversity term. The intervention nevertheless provides causal behavioural evidence that diversity modulation changes replacement activity under the tested conditions. The absence of a detectable final-error effect indicates that this behavioural change does not translate into a measurable performance difference at the tested budget. Accordingly, the result supports a causal effect on replacement behaviour, but not a claim that the diversity term is necessary for LAO or critical to its final optimization performance.

### E. Robustness and Cost

A local robustness/sensitivity analysis is reported using deterministic perturbations of five selected continuous search parameters at $\pm 10\%$, $\pm 20\%$, and $\pm 30\%$of their shipped defaults. The analysis uses 11 functions and 30 paired seeds, yielding 9,900 perturbation runs and 330 baseline/reference runs, for a total of 10,230 runs. The variance-based Sobol analysis was withdrawn because the attainable sample sizes failed the prespecified convergence check; the corresponding decision record is provided in the Supplementary Material. The five perturbed parameters are seasonal weight ($\alpha$), diversity weight ($\lambda$), local probability ($p_{\text{local}}$), step scale ($\sigma_0$), and phototropism ($\eta$). The environmental weight $\beta$, decay exponent $\gamma$, and discrete tournament size were not perturbed (see Limitations).

The median error ratio relative to the default does not exceed 1.021, with the worst case occurring for a 20% reduction in $p_{\text{local}}$. At the $\pm 30\%$level, the worst observed ratios are 1.011 for seasonal weight, 1.007 for diversity weight, 1.018 for local probability, 1.011 for step scale, and 1.002 for phototropism. Thus, across the five perturbed continuous parameters, the largest observed $\pm 30\%$perturbation is associated with an approximately 2% increase in median error. All 30 perturbation conditions are statistically detectable after Holm correction using paired Wilcoxon signed-rank tests on the seed-matched runs. The results indicate statistical sensitivity but small practical sensitivity over the tested perturbation range.

At equal evaluation budgets, the median wall-clock time per run at $D = 30$is 0.77 s for PSO, 1.10 s for GA, 2.24 s for WOA, 2.92 s for GWO, 3.00 s for DE, 3.34 s for LAO, 4.12 s for L-SHADE, 4.73 s for SHADE, and 10.53 s for CMA-ES. Because the protocol equalises the number of objective evaluations, these measurements do not establish evaluation efficiency. They instead characterise wall-clock overhead for the inexpensive CEC objectives used here. The $O(N^2D)$diversity computation contributes to LAO's computational overhead.

# VIII. DISCUSSION

## A. Interpretation of the Main Findings

Three main conclusions emerge from the present evidence. First, the final threshold-selection architecture is competitive under the tested constrained budget, attaining the third-best mean rank across the 29 benchmark functions at each tested dimension. This descriptive ranking should be distinguished from the pairwise inferential results, which do not establish a universal ordering among the algorithms. Second, the factorial ablation shows that the auxiliary mechanisms do not contribute equally: drift is harmful under the tested configuration, whereas the apparent benefit of elite refinement is no longer established after drift is removed. This conditional effect would be difficult to identify from a pooled analysis or a sequential leave-one-out design alone. Third, LAO's relative standing changes as the evaluation budget increases. This pattern is consistent with, but does not causally establish, an effect of the longer adaptation horizon available to success-history differential-evolution methods.

## B. Architectural Significance of Evaluation Gating

Beyond its empirical ranking, threshold-based selection makes explicit a distinction that is often implicit in evolutionary algorithm design: the point at which the selection decision enters the search cycle. Classical survivor selection operates after candidate generation and evaluation, whereas in the surrogate-assisted methods considered here, candidate construction precedes the decision about whether an exact evaluation is required. Threshold-based selection places the decision before candidate construction and therefore allows a failed threshold test to suppress both construction and exact evaluation. Definition 1 formalizes this distinction. This ordering is architectural rather than parametric. Because the gate precedes variation, an evaluation-gated mechanism is, in principle, compatible with different variation kernels without requiring changes to their internal update rules. CHC provides a close precedent through its pre-recombination threshold [37], but its threshold controls recombination rather than the coupled decision to construct and exactly evaluate a replacement. LAO places the gate earlier and couples threshold failure to both operations. Surrogate-assisted methods [39], [40] likewise reduce exact evaluations, but candidate construction has already occurred before the surrogate decision; this distinction may matter when candidate construction itself is computationally non-trivial, for example in constrained decoding or simulation-based proposal generation.

The practical significance of this distinction remains a hypothesis for expensive-objective settings. A single objective evaluation may require substantial simulation time in engineering design, simulation-based inference, or calibration of process-oriented models such as the phenology models discussed by Meier and Bigler [49]. The present experiments cannot test this computational advantage directly because the CEC 2017 objectives are inexpensive. What they establish is the architectural property itself: the algorithm can prevent a replacement from being constructed and evaluated when the threshold is not crossed.

## C. Behavioural Effects and Optimization Performance

The diversity intervention provides a clear separation between behavioural response and final optimization performance. Removing the diversity term significantly reduces late-run replacement activity ($p \approx 10^{-55}$ in the paired Wilcoxon analysis), but does not produce a detectable change in final error ($p = 0.31$ at $D = 10$and $p = 0.54$at $D = 30$). The intervention therefore supports a causal effect of diversity modulation on replacement behaviour under the tested conditions without establishing a corresponding performance effect. Two interpretations remain consistent with these results. First, the $300D$budget may be insufficient for a behavioural change of this magnitude to accumulate into a detectable difference in final error. Second, the diversity term may act through a pathway that is partly redundant with other components of the pressure equation. In particular, the seasonal schedule already concentrates replacement activity toward the latter part of the run, so removing the diversity contribution may alter the timing of replacement without materially changing the aggregate search pressure. The present intervention cannot distinguish between these explanations. The broader implication is methodological. A pre-specified behavioural prediction can be supported even when the corresponding performance prediction is not. In this case, the evidence justifies attributing a behavioural role to diversity modulation, but not treating the term as necessary for LAO or as a demonstrated source of improved final optimization performance.

## D. Implications for Algorithm-Design and the Biological Abstraction

The ablation results have implications beyond the particular LAO implementation. The original five-layer architecture was motivated by several biologically inspired mechanisms, yet the factorial analysis showed that their contributions were not equivalent: one component was harmful, while three others did not provide robust independent evidence of benefit under the tested protocol. This supports the use of component-level analysis in multi-mechanism optimizers, particularly where several plausible operators are introduced simultaneously. Factorial designs can also expose conditional effects that are difficult to identify through one-factor-at-a-time comparisons.

The biological mapping provides a further distinction. The phenological model contributes a quantitative seasonal structure through the Wang logistic formulation [3], whose timing and spread parameters retain interpretable links to leaf-fall observations. The wind, gust, and vibration mechanisms, although biologically motivated in the original design, did not survive the component analysis. The resulting design therefore treats biological inspiration as a hypothesis about mechanism rather than as a reason to retain an operator. This interpretation is consistent with the physiology literature [51], which distinguishes abscission from senescence as related but distinct processes, and supports the focus of the algorithm on the discrete abscission event rather than on the broader senescence cascade. The distinction is important for the interpretation of plant-inspired optimization more generally. A biological process can provide a useful quantitative structure without implying that the resulting optimizer reproduces the underlying biological system. In LAO, the value of the biological abstraction lies in the thresholded timing of replacement; its empirical relevance is ultimately determined by the behaviour and performance of the resulting optimization mechanism.

## IX. LIMITATIONS AND THREATS TO VALIDITY

The main empirical comparison is conducted at a budget of $300D$, which corresponds to approximately 3% of the $10{,}000D$ budget used in the CEC 2017 reference protocol [7]. The budget analysis shows that relative algorithmic performance changes with the available evaluation budget. The reported ranking results should therefore be interpreted within the budget regimes that were explicitly evaluated. The complete 29-function comparison at $10{,}000D$ was not completed for all three dimensions, while the four-budget analysis was conducted on a pre-specified stratified subset of 11 functions at $D = 10$. Full-suite confirmation at $1000D$ is available for $D = 10$ and $D = 30$; the corresponding $D = 50$ campaign is incomplete and is not included in full-suite inference. These additional experiments are used to characterise budget dependence and do not replace the primary $300D$ comparison.

The scope of the component-level conclusions is also limited. The factorial ablation was conducted at $D = 10$, whereas the diversity intervention covered $D = 10$ and $D = 30$. Consequently, component effects at higher dimensions should not be assumed without direct evidence. The theoretical analysis establishes an exact accounting identity for objective evaluations and an evaluation-indexed state formulation, but it does not establish convergence. In particular, the probability of global regeneration is state dependent, and the analysis does not provide a uniform positive lower bound. The literature review is also not a systematic survey; consequently, the novelty claim is restricted to the specific combination of rank-based strength, phenological scheduling, and per-individual threshold crossing and is stated as a qualified literature claim.

The baseline comparisons introduce additional protocol considerations. GWO and WOA use coefficient schedules reparameterized in terms of consumed evaluations rather than iteration count. CMA-ES uses disabled internal termination and a uniformly initialized mean rather than a fixed central initial mean. SHADE and L-SHADE use population sizes capped at 100 and $18D$, respectively, so that at least 50 generations fit within the reduced headline budget. These settings provide a common evaluation-budget protocol but are not exact reproductions of the canonical configurations used under the standard $10{,}000D$ budget. Accordingly, comparisons with these baselines should be interpreted within the experimental protocol used in this study.

Equal evaluation budgets also do not imply identical search trajectories. In particular, the initial population sizes differ across LAO ($N = 30$), SHADE ($N = 100$), and L-SHADE ($N = 18D$). Differences in population structure and generation count therefore remain relevant when interpreting results under a common NFE constraint. The sensitivity analysis covers five continuous parameters—$\alpha$, $\lambda$, $p_{\text{local}}$, $\sigma_0$, and $\eta$—while $\beta$, $\gamma$, and the discrete tournament size are not perturbed. The conclusions from this analysis are consequently local to the tested parameter neighbourhood.

All experiments were conducted under a single recorded software environment (CPython 3.13, NumPy 2.3, and pycma 4.4), within which 30 stored runs were reproduced deterministically. Reproducibility across different hardware, compilers, or BLAS backends was not evaluated. Finally, the CEC 2017 objectives used here are computationally inexpensive. The potential advantage of evaluation gating when exact objective evaluations dominate wall-clock cost is therefore a property implied by the architecture and supported by the broader expensive-black-box optimization setting, rather than a directly tested runtime effect in the present experiments.

## X. CONCLUSION

This paper formalised threshold-based selection as a mechanism in which the decision to initiate search is made before candidate construction and exact objective evaluation, and instantiated the mechanism through a quantitative abstraction of leaf abscission. Component-level analysis reduced the original five-layer design to a parsimonious core: drift was harmful under the tested configuration, while the remaining auxiliary components did not provide robust independent evidence of benefit after drift was removed. The resulting LAO algorithm remained competitive on the CEC 2017 suite under the tested equal-budget protocol at $D = 10, 30,$ and $50$, attaining the third-best mean rank among the nine compared algorithms. Its relative performance varied with the available evaluation budget, with the constrained-budget regime favouring LAO more strongly than the higher-budget regime. The present experiments do not establish an $FE/D$-only law or a statistically significant dimension-by-budget interaction.

The results also distinguish behavioural effects from final optimization performance. The factorial analysis showed that component effects can depend on interactions among mechanisms, while the diversity intervention demonstrated a causal change in replacement behaviour without a detectable change in final error at the tested budgets. These findings support evaluation gating as a distinct architectural design axis, but they do not imply universal superiority, evaluation efficiency in wall-clock terms, or performance benefits for every auxiliary component.

Future work should extend the budget analysis to complete full-suite comparisons across all dimensions, evaluate threshold-based selection against a broader range of restart, extinction, and population-management mechanisms, and examine adaptive success-history information as a possible extension of the threshold pressure. More generally, the results suggest that biologically motivated components in composite optimizers should be retained when their quantitative role is supported by component-level evidence rather than by biological plausibility alone.